\documentclass[letterpaper]{article} 
\usepackage{aaai2026}  
\usepackage{times}  
\usepackage{helvet}  
\usepackage{courier}  
\usepackage[hyphens]{url}  
\usepackage{graphicx} 
\usepackage{natbib}  
\usepackage{caption} 
\usepackage{algorithm}
\usepackage{algorithmic}

\usepackage{newfloat}
\usepackage{listings}

\DeclareCaptionStyle{ruled}{labelfont=normalfont,labelsep=colon,strut=off} 
\floatstyle{ruled}
\newfloat{listing}{tb}{lst}{}
\floatname{listing}{Listing}

\usepackage{amsmath}
\usepackage{amssymb}
\usepackage{mathtools}
\usepackage{amsthm}

\usepackage{booktabs} 
\usepackage{multirow}
\usepackage{makecell}
\usepackage{subcaption}
\usepackage{enumitem}
\usepackage{xcolor}

\nocopyright

\title{DW-DGAT: Dynamically Weighted Dual Graph Attention Network for Neurodegenerative Disease Diagnosis}
\author{
    Chengjia Liang\textsuperscript{\rm 1}, Zhenjiong Wang\textsuperscript{\rm 1}, Chao Chen\textsuperscript{\rm 2}, Ruizhi Zhang\textsuperscript{\rm 1}, Songxi Liang\textsuperscript{\rm 1},\\
    Hai Xie\textsuperscript{\rm 3}, Haijun Lei\textsuperscript{\rm 1}\thanks{Haijun Lei is the corresponding author.}, Zhongwei Huang\textsuperscript{\rm 4}
}
\affiliations{
    \textsuperscript{\rm 1}College of Computer Science and Software Technology, Shenzhen University, Shenzhen 518000, China\\

    \textsuperscript{\rm 2}School of Computer Science, University of Nottingham, Nottingham NG8 1BB, U.K.\\
    \textsuperscript{\rm 3}School of Cyberspace Security, Hainan University, Haikou 570228, China\\
    \textsuperscript{\rm 4}School of Computer Science, Hubei University of Technology, Wuhan 430068, China\\
    \{liangchengjia2022, 2021192003, 2022110191, 2022110319\}@email.szu.edu.cn, chao.chen@nottingham.ac.uk, xiehai@hainanu.edu.cn, lhj@szu.edu.cn, huangzhongwei@hbut.edu.cn
}

\usepackage{bibentry}

\begin{document}

\maketitle

\begin{abstract}
Parkinson\textquotesingle{}s disease (PD) and Alzheimer\textquotesingle{}s disease (AD) are the two most prevalent and incurable neurodegenerative diseases (NDs) worldwide, for which early diagnosis is critical to delay their progression. However, the high dimensionality of multi-metric data with diverse structural forms, the heterogeneity of neuroimaging and phenotypic data, and class imbalance collectively pose significant challenges to early ND diagnosis. To address these challenges, we propose a dynamically weighted dual graph attention network (DW-DGAT) that integrates: (1) a general-purpose data fusion strategy to merge three structural forms of multi-metric data; (2) a dual graph attention architecture based on brain regions and inter-sample relationships to extract both micro- and macro-level features; and (3) a class weight generation mechanism combined with two stable and effective loss functions to mitigate class imbalance. Rigorous experiments, based on the Parkinson Progression Marker Initiative (PPMI) and Alzheimer\textquotesingle{}s Disease Neuroimaging Initiative (ADNI) studies, demonstrate the state-of-the-art performance of our approach.
\end{abstract}

\begin{links}
    \link{Code}{https://github.com/AlexanderLeung9/DW-DGAT.git}
\end{links}

\section{Introduction}

Parkinson\textquotesingle{}s disease (PD) and Alzheimer\textquotesingle{}s disease (AD) are incurable neurodegenerative diseases (NDs) that primarily affect middle-aged and elderly individuals \cite{balestrino2020parkinson, scheltens2021alzheimer}. Over time, patients succumb to complications arising from NDs. Early diagnosis is crucial for initiating treatment to slow disease progression. However, neuroimaging changes in the early stages of NDs are often too subtle for physicians to detect. Specifically, there exists an intermediate prodromal (PRO) stage between healthy controls (HC) and PD, and an early mild cognitive impairment (EMCI) stage between cognitively normal (CN) and AD. The tiny differences among these cohorts necessitate the use of highly sensitive neuroimaging modalities for accurate diagnosis, among which magnetic resonance imaging (MRI) and diffusion tensor imaging (DTI) are two economical yet promising and complementary candidates.

T1-weighted MRI reflects static brain structures, including gray matter (GM), white matter (WM), and cerebrospinal fluid (CSF). DTI is an advanced MRI that can reveal pathological alterations in WM and has shown promise in distinguishing early NDs from healthy subjects \cite{poewe2017parkinson, mayo2019relationship}. However, fully exploiting DTI data remains challenging. First, DTI provides multiple 3D metrics, such as fractional anisotropy (FA), mean diffusivity (MD), local diffusion homogeneity based on Spearman's or Kendall's coefficient concordance (LDH-S or LDH-K) \cite{gong2013local}, the axial diffusivity (AXD, equivalent to the first eigenvalue L1), the radial diffusivity (RDD, defined as the mean of L2 and L3), and the 1st to 3rd eigenvectors (V1$\sim$V3), among others. Although these metrics offer complementary insights into the pathological brain changes, directly using all of them results in significant memory consumption and prohibitive computational costs. Second, when conducting brain region-level analysis on these metrics, it can yield 2D brain regional networks and 1D brain regional statistics that differ structurally from the original 3D metrics. As a result, existing studies often utilize only a limited subset of DTI or MRI metrics due to the difficulty of fusing multiple structural forms and high-dimensional data. For example, \citet{huang2025incomplete} employed GM from MRI along with L1 and V1 from DTI for PD diagnosis. \citet{song2022multicenter} combined resting-state functional MRI (rs-fMRI) and FA deterministic networks for AD diagnosis.

Regarding neuroimaging data, there are two main types of analysis methods. The first type aims to extract features directly from brain images but faces challenges such as high memory consumption and computational cost in high-dimensional 3D images \cite{cobbinah2022reducing, yang2023diagnosis} or incomplete perspectives in 2D slice images \cite{chen2024mvrna}. The second type segments the brain into multiple regions of interest (ROIs), constructs 2D brain networks, and applies graph-based analysis for feature extraction \cite{zhang2023graph, huang2024structural}. Therefore, it reduces the ND diagnosis problem to a graph classification task. Athough significant progress has been made, existing methods still rely on limited information sources, often overlooking other structural forms and data modalities.

In addition to neuroimaging data, phenotypic data, such as sex, age, and clinical assessment scores, are commonly used in diagnosis by physicians. However, directly fusing these low-dimensional but highly informative data with high-dimensional neuroimaging data often compromises their diagnostic value. This raises a challenge in designing an architecture capable of accurately extracting sample features at both levels. Graph neural networks (GNNs) have shown promise in modeling inter-sample relationships, but they face two key challenges. First, graph construction is critical, as there is no unique inherent graph that precisely reflects the objective relationships between samples. Second, GNNs typically perform better in transductive learning settings than in inductive ones; therefore, transductive learning is widely adopted in the neurological field \cite{ghorbani2022ra, song2022multicenter, zhang2022classification, liu2022mmgk, song2025knowledge}. However, transductive learning requires loading all samples during training, resulting in a significant memory burden and reduced flexibility for clinical use.

Finally, medical datasets often suffer from class imbalance, which leads conventional deep learning methods to misclassify minority-class samples as majority-class ones. Traditional strategies typically rely on over-sampling via data generation \cite{balakrishnan2023mahalanobis} based on minority classes, or on under-sampling via a bootstrapping method \cite{roy2023imbalanced} or feature fusion \cite{yue2024specificity} based on majority classes. However, data generation is complex for multi-metric and multi-form medical data that contains numerous subtle yet critical details; meanwhile, under-sampling risks severe overfitting in small-sized, high-dimensional datasets. Hence, cost-sensitive approaches have emerged as promising alternatives in medical image analysis \cite{kavitha2024enhanced}. However, a single fixed cost function tailored to a specific dataset structure often fails to generalize to more diverse datasets.

In this study, we fuse MRI and DTI metrics of different structural forms through a general-purpose and efficient strategy to maximize their utility in ND diagnosis. Based on the fused data, we design a single graph attention (SGA) module to learn long-range dependencies among ROIs, and a global graph attention (GGA) module based on phenotypic data to capture demographic relationships among subjects. Lastly, we incorporate a class weight generator (CWG) that dynamically generates class weights to guide the classifier within an cooperative training framework.

Our contributions are summarized as follows.

\begin{itemize}
    \item We propose a data fusion (DF) method to efficiently fuse 1D, 2D and 3D multi-metric data for subsequent analysis.
    \item We design a dual graph attention network (DGAT) to learn micro- and macro-level graph features.
    \item We develop a dynamically weighted (DW) mechanism to alleviate the issue of class imbalance in medical datasets.
    \item The superiority and generalizability of the proposed DW-DGAT are validated on PD and AD datasets.
\end{itemize}

\section{Related Work}
\subsubsection{Data fusion.}
Although abundant multi-metric data can provide rich information for deep neural networks, efficiently fusing data with diverse structural forms for diagnosis remains challenging. For instance, some studies integrate multiple structural forms of data in a high-level embedding space using specialized architectures \cite{bi2024gan, ding2024parkinson}, but these designs are typically limited to a few specific metrics and structural forms. Recently, \citet{bourriez2024chada} introduced the ChAda-ViT network, which adaptively handles images with varying numbers of channels and projects them into a common embedding space. However, this method is restricted to 2D image data. A generalized approach capable of integrating multi-metric data across cubic, planar, and linear structures is still lacking.

\subsubsection{Brain network analysis.}
Since the brain can be viewed as an ensemble of ROIs, many researchers have focused on analyzing graphs constructed from these regions and their interconnections. Convolutional layers and fully connected (FC) layers are widely used for message passing in ND-specific GNNs, aggregating information from node neighbors layer by layer \cite{kawahara2017brainnetcnn, xu2021graph, kan2022fbnetgen, zhou2024lcgnet}, thereby enabling feature extraction from a single graph. Another important analytical technique is the attention mechanism \cite{kan2022brain, ying2021transformers, yu2024long, shehzad2025multiscale}, which encodes nodes and edges via multi-head self-attention (MHSA) or graph attention networks (GATs), capturing global dependencies among ROIs. Additionally, graph pooling is widely employed in neuroscience to select key ROIs while discarding irrelevant ones, thus enabling versatile GNNs to adapt to brain networks \cite{peng2020motif, li2021braingnn, zhang2023gcl, xu2024contrastive}.

\subsubsection{Subject relationship analysis.}
Different from brain network analysis, subject relationship analysis addresses the problem of node prediction and primarily focuses on phenotypic data. There are mainly two types of methods to leverage phenotypic data. The first type incorporates phenotypic data without any learnable parameters, preserving their original characteristics \cite{ghorbani2022ra, liu2022mmgk}. Consequently, the diagnostic performance heavily depends on the quality of the phenotypic data used to construct a graph representing subject relationships. In contrast, the second type of methods incorporates learnable parameters to model phenotypic data, aiming to construct a more accurate relationship graph among subjects \cite{song2022multicenter, zhang2022classification, song2025knowledge}. However, these methods risk constructing suboptimal graphs when noisy node features are included in the adjacency graph. Recently, \citet{chen2024gcn} proposed GCN-MHSA, which builds the adjacency graph solely from node features via an attention mechanism, in contrast to previous methods that rely on convolutional or affine operations.

\subsubsection{Class imbalance.}
Medical datasets often suffer from the issue of class imbalance. To alleviate this, \citet{ghorbani2022ra} proposed RA-GCN, a network based on a re-weighted cost-sensitive strategy that resembles the training paradigm of generative adversarial networks (GANs) \cite{goodfellow2020generative}. However, GAN training is notoriously unstable and prone to collapse without appropriate hyperparameter tuning and regularization \cite{arjovsky2017towards}. Although several techniques have been proposed to address these issues \cite{gulrajani2017improved}, RA-GCN remains a relatively plain design. In our experiments, it occasionally exhibits instability and fails to converge. In this work, we propose a similar generative cooperative strategy but go a step further by explicitly addressing the instability and non-convergence issues inherent in GAN training.

\begin{figure*}[t]
    \begin{center}
        \centerline{\includegraphics[width=\textwidth]{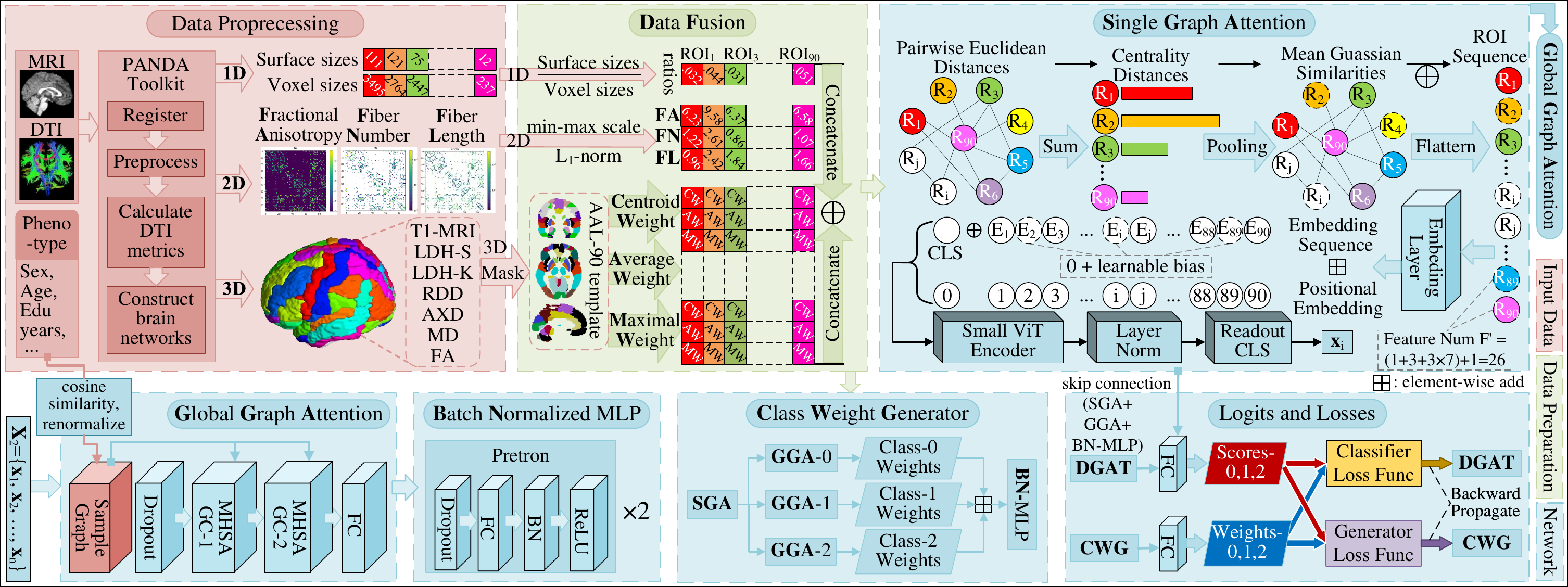}} 
        \caption{The architecture of the proposed methodology.}
        \label{fig1}
    \end{center}
\end{figure*}

\section{Datasets}
We built two datasets from the Parkinson Progression Marker Initiative (PPMI) \cite{marek2018parkinson} and the Alzheimer\textquotesingle{}s Disease Neuroimaging Initiative 3 (ADNI3) \cite{weiner2017alzheimer}, which are globally authoritative and publicly available ND data sources. The PPMI dataset contains three cohorts: HC, PRO, and PD, with 69, 72, and 175 subjects, and 121, 123, and 392 neuroimaging samples, respectively. The ADNI3 dataset contains another three cohorts: CN, EMCI, and AD, with 163, 118, and 29 subjects, and 234, 193, and 37 neuroimaging samples, respectively. All acquisitions are at time points of 0, 12, and 24 months.

We used the PANDA toolkit \cite{cui2013panda}, which provides a fully automated pipeline based on the FMRIB Software Library (FSL) \cite{smith2004advances}, to preprocess the data. As shown in Figure~\ref{fig1}, the pipeline first registers DTI to T1-MRI, and then generates six 3D diffusion metrics with a voxel size of 2 mm in standard space. Furthermore, three 2D deterministic brain region connectivity networks—FA, FN (Fiber Number), and FL (Fiber Length)—are constructed, along with two 1D vectors containing statistical information on ROI surface size and ROI voxel size. Sex, age, years of education, race, and Montreal Cognitive Assessment (MoCA) scores are used as phenotypes in both datasets. Additionally, the PPMI dataset includes MDS-UPDRS Part II scores, and the ADNI3 dataset includes Mini-Mental State Examination (MMSE) scores. These three clinical assessment scores were selected because they are easy to administer by community health providers and patients.

\section{Methodology}
The overall architecture of our proposed methodology is illustrated in Figure~\ref{fig1}, which consists of four key components: DF, SGA, GGA, and CWG. The DF module fuses the 1D, 2D, and 3D data into a matrix $\mathbf{X}$ whose rows represent ROIs and columns correspond to metrics. The SGA first constructs an ROI graph from $\mathbf{X}$, and then pools the graph and flattens it back into an ROI sequence. Next, it uses a small ViT encoder \cite{dosovitskiy2020image} to compute the class token (CLS) for each sample. The GGA dynamically constructs a phenotypic adjacency graph based on the current batch of samples and captures sample-level graph features through two MHSA graph convolution (GC) layers. In the training phase, the CWG dynamically adjusts the class weights of each sample to balance the classifier output.

\subsection{Data Fusion}
This module fuses three structural forms of metrics. Let $R=90$ be the ROI count based on the AAL-90 atlas. To merge the two 1D vectors, we compute the ratio of the number of surface voxels to the total voxel count in each ROI. For each 2D deterministic network, we first rescale all values to the range $[0,1]$ using a min-max function, and then apply the $L_1$-norm to each row to produce an $R$-dimensional vector. Let the 3D neuroimage of every metric be $\{w_{i,j,k} | i,j,k \in \mathbb{N}, w_{i,j,k} \geq 0, w_{i,j,k} \in \mathbb{R}\}$, and the brain template be $\{t_{i,j,k} | t_{i,j,k} \in [1, R], i,j,k,t_{i,j,k} \in \mathbb{N}\}$. We segment the ROIs by masking $w_{i,j,k}$ with $t_{i,j,k}$ along the ROI numbers $\{r|r \in [1, R], r \in \mathbb{N}\}$. Specifically, for $i,j,k$ that satisfy $t_{i,,j,k}=r$, the corresponding $w_{i,j,k}$ are collected. For each ROI, we calculate its centroid coordinates $\left(\bar{x},\ \bar{y},\ \bar{z}\right)_r$, centroid weight $w_r$, average weight ${\bar{w}}_r$, and maximum weight ${\hat{w}}_r$, as defined in Eq. \ref{eq1}, \ref{eq2}, \ref{eq3}, \ref{eq4}.

\begin{equation}
    \label{eq1}
    \begin{split}
        (\bar{x}, \bar{y}, \bar{z})_{r} &= \frac{1}{\sum_{i, j, k} w_{i, j, k}} * \\
        &\left(\sum_{i, j, k} i \cdot w_{i, j, k}, \sum_{i, j, k} j \cdot w_{i, j, k}, \sum_{i, j, k} k \cdot w_{i, j, k}\right), \\
        &i, j, k \in\left\{i, j, k \mid t_{i, j, k}=r\right\}
    \end{split}
\end{equation}

\begin{equation}
    \label{eq2}
    w_r=w_{\text {round }(\bar{x}), \text { round }(\bar{y}), \text { round } (\bar{z})} 
\end{equation}

\begin{equation}
    \label{eq3}
    \bar{w}_{r}=\frac{\sum_{i, j, k} w_{i, j, k}}{\left| \left\{w_{i, j, k}\right\} \right|}, \quad i, j, k \in \left\{i, j, k \mid t_{i, j, k}=r\right\}
\end{equation}

\begin{equation}
    \label{eq4}
    \hat{w}_{r}=\max \left(\left\{w_{i, j, k}\right\}\right), i, j, k \in\left\{i, j, k \mid t_{i, j, k}=r\right\}
\end{equation}

\noindent Since the MRI-derived features exceed 1 in range, they are normalized by dividing both $\mathrm{w}_{r}$ and $\bar{w}_{r}$ by $\hat{w}_{r}$, and scaling each $\hat{w}_{r}$ by $\max\left(\{\hat{w}_{r}\}\right)$. Next, we concatenate $w_r$, ${\bar{w}}_r$, and ${\hat{w}}_r$ to form a vector. Finally, we concatenate the 1D, 2D, and 3D features for each ROI, and then stack the ROI features into $\mathbf{X} \in \mathbb{R}^{R \times F}$ for each sample.

\subsection{Single Graph Attention}
The SGA module utilizes two crucial techniques to extract features of an ROI graph.

The first is graph pooling. Initially, a matrix is computed based on the pairwise Euclidean distances between the ROI features. The centrality distance of each ROI is then obtained by summing its distances to all other ROIs. Subsequently, we zero out the features for the 50\% of ROIs with the greatest centrality distances, as these ROIs are assumed to have weaker connectivity and thus contribute less to ND diagnosis. Furthermore, a Gaussian kernel function is used to compute the mean similarity of each ROI based on its centrality distance, with the kernel width $\sigma$ set to the average of $R$ centrality distances. After appending the mean similarity to the metric features of each ROI, all ROIs are stacked into $\mathbf{X}_1 \in \mathbb{R}^{R \times F'}$ according to their original indexing order.

The second technique is the ViT encoder. SGA uses an embedding layer to project $\mathbf{X}_1$ into $\mathbf{Y}_1 \in \mathbb{R}^{R \times E}$, where $E=384$. A learnable CLS $\mathbf{t}_1 \in \mathbb{R}^{1 \times E}$ is then prepended to $\mathbf{Y}_1$, yielding the extended ROI embeddings $\mathbf{Y}_2 \in \mathbb{R}^{(1+R) \times E}$. Next, learnable positional embeddings $\mathbf{P}_2 \in \mathbb{R}^{(1+R) \times E}$ are added to $\mathbf{Y}_2$, resulting in $\mathbf{Y}_3$, which enables SGA to capture positional relationships among ROIs. $\mathbf{Y}_3$ is fed into a small ViT encoder composed of $12$ MHSA blocks to compute global attention. Layer normalization is applied afterward to stabilize the output. Finally, the CLS $\mathbf{x}_i$ is read out as the representation of the sample. $N$ sample representations form the sample matrix $\mathbf{X}_2 = [\mathbf{x}_1, \mathbf{x}_2, \ldots, \mathbf{x}_N]^\top \in \mathbb{R}^{N \times E}$.

\subsection{Global Graph Attention}
We observed an important phenomenon in our experiments: conventional Graph Convolutional Network (GCN) tends to perform better when the underlying adjacency graph more accurately reflects the true relationships among samples. To this end, we enhance the GCN from two aspects.

First, we construct the adjacency matrix $\mathbf{A}_2$ from the static phenotypic feature matrix $\mathbf{P}=[\mathbf{p}_1, \mathbf{p}_2, \ldots, \mathbf{p}_N]^\top \in \mathbb{R}^{N \times P}$. Specifically, we measure the pairwise distances between samples based on the transformed cosine similarity defined in Eq. \ref{eq5}. Then, we apply a Gaussian kernel function, with $\sigma$ set to the median of all off-diagonal pairwise distances, to build a similarity matrix $\mathbf{A}_1\in\mathbb{R}^{N \times N}$. Next, we remove self-loops from $\mathbf{A}_1$ and apply the renormalization trick \cite{kipf2016semi} to obtain $\mathbf{A}_2$.

\begin{equation}
    \label{eq5}
    d_{i, j}=\left\{
    \begin{array}{ll}
        1-\frac{\mathbf{p}_{i} \cdot \mathbf{p}_{j}}{\left\|\mathbf{p}_{i}\right\|_{2} \cdot\left\|\mathbf{p}_{j}\right\|_{2}}, & i \neq j \\
        1, & i=j
    \end{array}\right.
\end{equation}

Second, we redesign the GC layer of the GCN by replacing its affine weight matrix with MHSA. Let $(\mathbf{H}_k)_{i,j}$ be the element in the $i$-th row and $j$-th column of a matrix. We compute the matrix $\mathbf{H}_k \in \mathbb{R}^{N \times E}$ for the $k$-th sample as shown in Eq. \ref{eq6}.

\begin{equation}
    \label{eq6}
    (\mathbf{H}_k)_{i,j} = (\mathbf{A}_2)_{k,i} * (\mathbf{X}_2)_{i,j}
\end{equation}

\noindent Then, we compute the neighbor aggregation $\mathbf{Z}_k^h \in \mathbb{R}^{N \times d}$ for each attention head to adaptively adjust the edge weights of the adjacency graph, as defined in Eq.~\ref{eq7} and Eq.~\ref{eq8}, where $h \in [1, H]$, $\mathbf{W}_Q, \mathbf{W}_K, \mathbf{W}_V \in \mathbb{R}^{E \times E}$, and $d = E/H$. We first obtain $\mathbf{Q}_k, \mathbf{K}_k, \mathbf{V}_k \in \mathbb{R}^{N \times E}$ and then split them along the feature dimension into $H$ heads, i.e., $\mathbf{Q}_k=[\mathbf{Q}_k^1, \mathbf{Q}_k^2, \ldots, \mathbf{Q}_k^H]$ with $\mathbf{Q}_k^h \in \mathbb{R}^{N \times d}$. The per-head outputs are concatenated as $\mathbf{Z}_k = [\mathbf{Z}_k^1, \mathbf{Z}_k^2, \ldots, \mathbf{Z}_k^H] \in \mathbb{R}^{N \times E}$, and the outputs from all samples are stacked to form $\mathbf{Z} = [\mathbf{Z}_1; \mathbf{Z}_2; \ldots; \mathbf{Z}_N] \in \mathbb{R}^{N \times N \times E}$.

\begin{equation}
    \label{eq7}
    \mathbf{Z}_k^h = \text{softmax}(\frac{\mathbf{Q}_k^h (\mathbf{K}_k^h)^\top}{\sqrt{d}}) \mathbf{V}_k^h
\end{equation}

\begin{equation}
    \label{eq8}
    \mathbf{Q}_k, \mathbf{K}_k, \mathbf{V}_k = \mathbf{H}_k * (\mathbf{W}_Q, \mathbf{W}_K, \mathbf{W}_V)
\end{equation}

\noindent Finally, the hidden representation of the GC layer is obtained by summing $\mathbf{Z}$ along the second axis, i.e., $\mathbf{H} = \sum_{j} \mathbf{Z}_{i,j,k} \in \mathbb{R}^{N \times E}$, followed by a layer normalization and a GELU activation \cite{hendrycks2016gaussian}. In addition, each of the two MHSA-GC layers outputs features with twice the dimensionality of its input, while the final FC layer reduces the dimensionality to half of its input.

\subsection{Class Weight Generator}
The architecture of CWG is highly similar to DGAT, except that it consists of $C$ GGAs, where $C$ denotes the number of classes. The supporting graphs of GGAs are constructed by masking $\mathbf{A}_2$ according to the corresponding class labels, allowing each GGA to focus exclusively on the samples within its class. The logits generated by the GGAs are summed element-wise, and then passed through a batch normalized (BN) MLP to produce the final class weights.

Compared to RA-GCN \cite{ghorbani2022ra}, we develop two more stable and effective loss functions for both the generator and the classifier. The class weights are denoted as $\mathbf{W}=[\mathbf{w}_1, \mathbf{w}_2, \ldots, \mathbf{w}_N]^\top \in \mathbb{R}^{N \times C}$; the DGAT scores as $\mathbf{S}=[\mathbf{s}_1, \mathbf{s}_2, \ldots, \mathbf{s}_N]^\top \in \mathbb{R}^{N \times C}$; and the one-hot labels as $\mathbf{Y} \in \mathbb{N}^{N \times C}$. We define the two losses $L_1$ in Eq.~\ref{eq9} for DGAT and $L_3$ in Eq. \ref{eq14} for CWG, where $\operatorname{CE}_\text{SGA}$ is a standard cross-entropy loss computed from the SGA logits and $\mathbf{Y}$, and $\alpha = 0.5$ is a hyperparameter.

\begin{equation}
    \label{eq9}
    L_1 = \min_\text{DGAT} \frac{1}{2} \left(L_2 + \operatorname{CE}_\text{SGA} \right)
\end{equation}

\begin{equation}
    \label{eq10}
    L_2 = -\frac{1}{N} \sum_{i=1}^{N} \sum_{j=1}^{C} \mathbf{Y}_{i,j} \cdot \mathbf{R}_{i,j} \cdot \log \mathbf{O}_{i,j}
\end{equation}

\begin{equation}
    \label{eq11}
    \mathbf{O}_{i,j} = \frac{\exp\left(\mathbf{S}_{i,j} - \max \left(\mathbf{s}_i\right)\right)}{\sum_{k=1}^{C} \exp\left(\mathbf{S}_{i,k} - \max \left(\mathbf{s}_i\right)\right)}
\end{equation}

\begin{equation}
    \label{eq12}
    \mathbf{R}_{i,j} = \frac{\mathbf{Q}_{i,j}}{\min \left(\mathbf{q}_i\right)}
\end{equation}

\begin{equation}
    \label{eq13}
    \mathbf{Q}_{i,j} = \frac{\exp\left(-\left(\mathbf{W}_{i,j} - \min\left(\mathbf{w}_i\right)\right)\right)}{\sum_{k=1}^{C} \exp\left(-\left(\mathbf{W}_{i,k} - \min\left(\mathbf{w}_i\right)\right)\right)} + \epsilon
\end{equation}

\begin{equation}
    \label{eq14}
    L_3 = \min_\text{CWG} \left( L_2 - \frac{\alpha}{N \cdot C} \sum_{i=1}^{N} \sum_{j=1}^{C} \mathbf{R}_{i,j} \cdot \log \mathbf{R}_{i,j} \right)
\end{equation}

Eq.~\ref{eq11} and Eq.~\ref{eq13} subtract the extreme elements of vectors to align their values on the same side of the y-axis, thus avoiding division by large numbers in the softmax function, which may cause numerical instability. A small constant $\epsilon$ equal to the double-precision machine epsilon is added to Eq.~\ref{eq13} to prevent gradient vanishing. To address class imbalance, Eq.~\ref{eq13} inverts the numerical relationship among class weights, while Eq.~\ref{eq12}  penalizes the classifier for incorrect class predictions without affecting the other classes, thus ensuring function stability. Algorithm~\ref{alg1} outlines the training steps in an epoch for CWG and DGAT.

\begin{algorithm}[tb]
    \caption{Training procedure of CWG and DGAT.}
    \label{alg1}
    \textbf{Input}: $\mathbf{X}_1'=[\mathbf{X}_1^1; \mathbf{X}_1^2; \ldots; \mathbf{X}_1^N] \in \mathbb{R}^{N \times R \times F'}$, $\mathbf{P}$, $\mathbf{Y}$\\
    \textbf{Output}: DGAT
    \begin{algorithmic}[1]
        \FOR{a batch of samples in all training samples}
            \STATE Build the adjacency graph $\mathbf{A}_2$ from $\mathbf{P}$
            \STATE Build masked graphs $\mathbf{M}_0$, $\mathbf{M}_1$, $\mathbf{M}_2$ from $\mathbf{Y}$
            \STATE $\mathbf{A}_2^i = \mathbf{A}_2 * \mathbf{M}_i, i \in [0, C), i \in \mathbb{N}$
            \STATE $\mathbf{W}_1^i = \text{CWG}_i\text{.forward}(\mathbf{X}_1', \mathbf{A}_2^i) \in \mathbb{R}^{N \times C}$
            \STATE $\mathbf{W}_1 = [{\mathbf{W}_1^i[:, i]}]_{i=0}^{C-1} \in \mathbb{R}^{N \times C}$~~~~// concat vectors
            \STATE $\mathbf{S} = \text{DGAT.forward}(\mathbf{X}'_1, \mathbf{A}_2) \in \mathbb{R}^{N \times C}$
            \STATE $L_3 = \text{CWG.cooperative\_loss}(\mathbf{W}_1, \mathbf{Y}, \mathbf{S})$
            \STATE Descend gradients of CWG from $L_3$
            \STATE Update parameters of CWG
            \STATE $\mathbf{W}_2^i = \text{CWG}_i\text{.forward}(\mathbf{X}_1', \mathbf{A}_2^i),~i \in \{0, 1, 2\}$
            \STATE $\mathbf{W}_2 = [{\mathbf{W}_2^i[:, i]}]_{i=0}^{C-1} \in \mathbb{R}^{N \times C}$~~~~// class weights
            \STATE $L_1 = \text{DGAT.cooperative\_loss}(\mathbf{S}, \mathbf{Y}, \mathbf{W}_2)$
            \STATE Descend gradients of DGAT from $L_1$
            \STATE Update parameters of DGAT
        \ENDFOR
        \STATE {\bfseries return} DGAT
    \end{algorithmic}
\end{algorithm}

\section{Experiments}
\subsection{Implementation Details}
The proposed method was implemented using Python 3.9, PyTorch 1.13.1+cu116, and CUDA drivers of v525 or v535 based on Ubuntu 20.04. All networks ran on a 24 GB GPU. The random seed was fixed at 231 for all experiments. GGA used $H = 6$ on the PPMI dataset and $H = 8$ on the ADNI3 dataset. Our network was trained for 500 epochs using the Adam optimizer \cite{kingma2014adam} with a learning rate of 0.001. The dropout rate was set to 0.5. The batch size was set to 64 for both our network and baseline networks. All experiments were evaluated via ten-fold cross-validation (CV). To prevent data leakage, samples from the same subject at different time points were strictly assigned to either the training set or the test set.

The evaluation metrics include accuracy (ACC), balanced accuracy (BA), F1 score, and specificity (SPE). BA accounts for class imbalance, where 1.0 indicates perfect classification, and $\frac{1}{C}$ or lower values mean a random guess.

\subsection{Comparative Experiments}
We compared our method with three categories of networks across the two datasets. These include public vision networks (a three-layer MLP, VGG-19-BN \cite{simonyan2014very}, ResNet-34 \cite{he2016deep}, and a small ViT \cite{dosovitskiy2020image}), public GNNs (GCN \cite{kipf2016semi}, MA-GCNN \cite{peng2020motif}, GATv2 \cite{brody2021attentive}, ChebNetII \cite{he2022convolutional}, and GCN-MHSA \cite{chen2024gcn}), and ND-specific methods (BrainNetCNN \cite{kawahara2017brainnetcnn}, BrainGNN \cite{li2021braingnn}, BrainNet-T \cite{kan2022brain}, LG-GNN \cite{zhang2022classification}, and RA-GCN \cite{ghorbani2022ra}). Since the comparison networks cannot handle the full high-dimensional input, we provided them with core metrics—three 2D deterministic matrices—to ensure compatibility. Specifically, networks based on 2D input received three vertically stacked matrices. For 1D-input networks, we followed the practices of \citet{song2022multicenter} and \citet{huang2024structural}, i.e., extracting the upper triangular elements of each symmetric matrix and then concatenating them into a 1D vector. The phenotypic graph was adopted for all public GNNs that rely on an adjacency graph but do not define a specific construction method. The input formats and other settings of ND-specific methods were kept as their original design. All models were trained until they approximately reached their minimum loss, with training capped at 500 epochs. Finally, we reported the average of their best evaluation performance across all folds, as presented in Table~\ref{tab1}.

\begin{table*}[t]
    \begin{center}
    \begin{small}
    \begin{tabular}{lcccccccc}
        \toprule
        \multirow{2}{*}[-0.5ex]{\makecell[l]{Methods or\\ Networks}} & \multicolumn{4}{c}{\textbf{HC \textit{vs.} PRO \textit{vs.} PD}} & \multicolumn{4}{c}{\textbf{CN \textit{vs.} EMCI \textit{vs.} AD}} \\
        \cmidrule(lr){2-5} \cmidrule(lr){6-9}
        ~ & ACC(\%) & BA(\%) & F1(\%) & SPE(\%) & ACC(\%) & BA(\%) & F1(\%) & SPE(\%) \\
        \hline
        BrainGNN & 64.15±3.96 & 41.02±5.45 & 54.83±6.56 & 70.51±2.72 & 56.90±3.10 & 44.15±8.19 & 53.76±6.67 & 72.08±4.10 \\
        BrainNet-T & 63.52±3.47 & 38.10±5.44 & 53.11±8.53 & 69.05±2.72 & 56.47±3.29 & 40.86±2.57 & 54.04±2.57 & 70.43±1.29 \\
        BrainNetCNN & 61.79±3.09 & 37.92±6.31 & 53.01±9.11 & 68.96±3.15 & 53.66±8.73 & 38.13±7.16 & 50.95±8.12 & 69.07±3.58 \\
        ChebNetII & 66.99±4.13 & 44.01±5.74 & 58.17±5.57 & 72.00±2.87 & 62.90±2.76 & \underline{50.15±8.34} & 58.36±5.57 & \underline{75.07±4.17} \\
        GATv2 & 65.08±3.52 & 40.98±7.06 & 54.79±5.72 & 70.49±3.53 & 62.07±2.82 & 48.39±4.95 & 58.86±5.49 & 74.19±2.48 \\
        GCN & 63.96±3.93 & 38.97±4.12 & 53.60±5.57 & 69.49±2.06 & 59.41±3.97 & 41.29±4.78 & 53.50±7.74 & 70.64±2.39 \\
        GCN-MHSA & 61.54±4.21 & 33.33±0.00 & 46.97±5.18 & 66.67±0.00 & 50.86±3.22 & 34.19±2.75 & 40.03±4.83 & 67.10±1.38 \\
        LG-GNN & 66.04±4.80 & 42.61±5.59 & 56.92±7.44 & 71.30±2.80 & 61.64±3.56 & 43.56±3.36 & 58.08±3.88 & 71.78±1.68 \\
        MA-GCNN & 50.69±12.23 & 37.50±3.96 & 44.12±13.46 & 68.75±1.98 & 53.28±3.49 & 36.64±3.71 & 40.45±7.07 & 68.32±1.85 \\
        MLP & 63.26±3.98 & 39.08±5.26 & 52.78±5.53 & 69.54±2.63 & 56.69±4.84 & 40.08±5.60 & 47.67±10.45 & 70.04±2.80 \\
        RA-GCN & 61.54±4.21 & 33.33±0.00 & 46.97±5.18 & 66.67±0.00 & 55.07±3.92 & 37.02±4.30 & 44.12±9.12 & 68.51±2.15 \\
        ResNet-34 & 62.57±3.84 & 39.20±3.76 & 53.74±5.83 & 69.60±1.88 & 54.42±4.78 & 40.56±4.77 & 49.03±7.30 & 70.28±2.39 \\
        VGG-19-BN & 63.96±2.81 & 38.61±3.73 & 53.51±2.55 & 69.31±1.87 & 60.64±4.28 & 42.41±3.70 & 55.93±5.21 & 71.21±1.85 \\
        \underline{ViT-small} & \underline{66.99±3.23} & \underline{45.65±5.53} & \underline{59.54±4.61} & \underline{72.82±2.77} & \underline{64.03±3.93} & 46.71±6.23 & \underline{59.62±6.52} & 73.35±3.11 \\
        \textbf{DW-DGAT} & \textbf{74.56±5.99} & \textbf{59.31±8.73} & \textbf{70.57±7.31} & \textbf{79.66±4.36}& \textbf{68.65±4.35} & \textbf{66.18±9.48}& \textbf{66.79±5.23} & \textbf{83.09±4.74} \\
        \bottomrule
    \end{tabular}
    \end{small}
    \end{center}
    \caption{Evaluation scores of all methods and networks based on ten-fold cross-validation. Bold font indicates the best scores, while underlined font indicates the second-best.}
    \label{tab1}
\end{table*}

\begin{table*}[t]
    \begin{center}
    \begin{small}
    \begin{tabular}{lcccccccc}
        \toprule
        \multirow{2}{*}[-0.5ex]{\makecell[l]{Modules}} & \multicolumn{4}{c}{\textbf{HC \textit{vs.} PRO \textit{vs.} PD}} & \multicolumn{4}{c}{\textbf{CN \textit{vs.} EMCI \textit{vs.} AD}} \\
        \cmidrule(lr){2-5} \cmidrule(lr){6-9}
        ~ & ACC(\%) & BA(\%) & F1(\%) & SPE(\%) & ACC(\%) & BA(\%) & F1(\%) & SPE(\%) \\
        \hline
        baseline & 63.05±3.62 & 36.15±5.09 & 50.64±6.86 & 68.08±2.54 & 56.03±3.23 & 40.58±4.46 & 52.13±6.50 & 70.29±2.23 \\
        DF & 65.41±3.84 & 44.18±6.97 & 58.78±9.13 & 72.09±3.48 & 60.78±3.94 & 43.90±5.05 & 58.11±6.74 & 71.95±2.53 \\
        DF+SGA & 67.45±3.63 & 46.62±4.25 & 61.50±5.45 & 73.31±2.12 & 61.42±4.29 & 47.27±7.35 & 59.48±9.80 & 73.63±3.67 \\
        DF+SGA+GGA & 71.70±5.30 & 55.59±9.58 & 68.42±10.16 & 77.79±4.79 & 65.09±3.14 & 64.31±9.97 & 64.66±10.44 & 82.15±4.99 \\
        \textbf{complete} & \textbf{74.56±5.99} & \textbf{59.31±8.73} & \textbf{70.57±7.31} & \textbf{79.66±4.36}& \textbf{68.65±4.35} & \textbf{66.18±9.48}& \textbf{66.79±5.23} & \textbf{83.09±4.74} \\
        \bottomrule
    \end{tabular}
    \end{small}
    \end{center}
    \caption{Ablation experiments on DW-DGAT. The baseline is a basic two-layer MLP based on three deterministic networks.}
    \label{tab2}
\end{table*}

\begin{figure}[tb]
    \begin{center}
    \centerline{\includegraphics[width=\columnwidth]{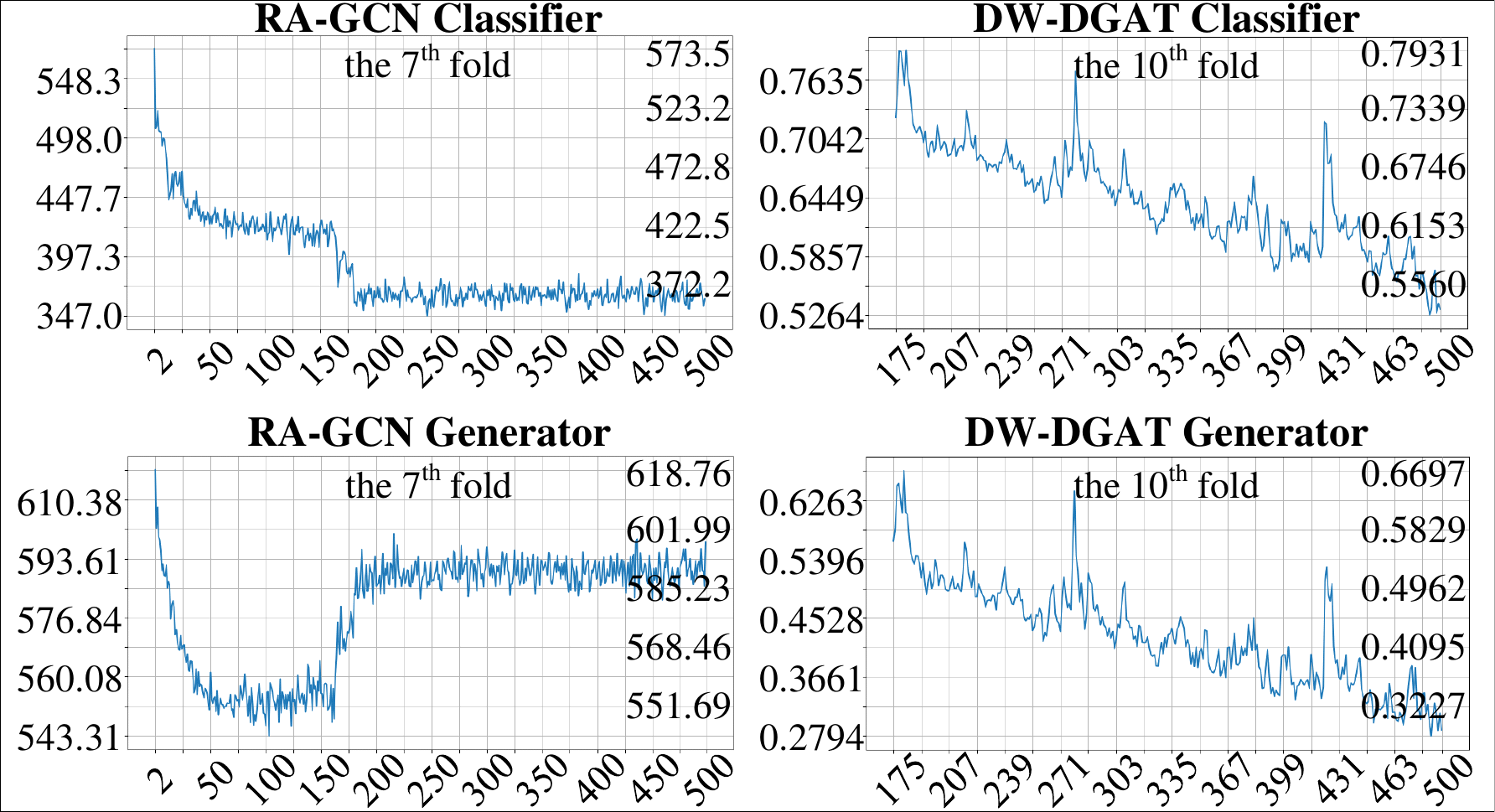}}
    \caption{Loss histories of RA-GCN and DW-DGAT.}
    \label{fig2}
    \end{center}
\end{figure}

From Table \ref{tab1}, we observe that our method outperforms the second-best network by 7.57\% on HC \textit{vs.} PRO \textit{vs.} PD and by 4.62\% on CN \textit{vs.} EMCI \textit{vs.} AD in terms of accuracy. The best-performing vision network, GNN, and ND method are ViT-small, ChebNetII, and LG-GNN, respectively. Among them, ViT-small and ChebNetII achieve comparable performance, both surpassing LG-GNN. Notably, RA-GCN suffers from overfitting on HC \textit{vs.} PRO \textit{vs.} PD. To further investigate this, we compared the training loss histories of RA-GCN and DW-DGAT in their best-performing folds, as illustrated in Figure~\ref{fig2}. The figure reveals that the adversarial training architecture of RA-GCN hampers effective convergence, with its loss stagnating at a high level. In contrast, the loss of DW-DGAT gradually decreases, demonstrating the stability and effectiveness of the proposed CWG.

\begin{figure}[tb]
    \begin{center}
    \centerline{\includegraphics[width=\columnwidth]{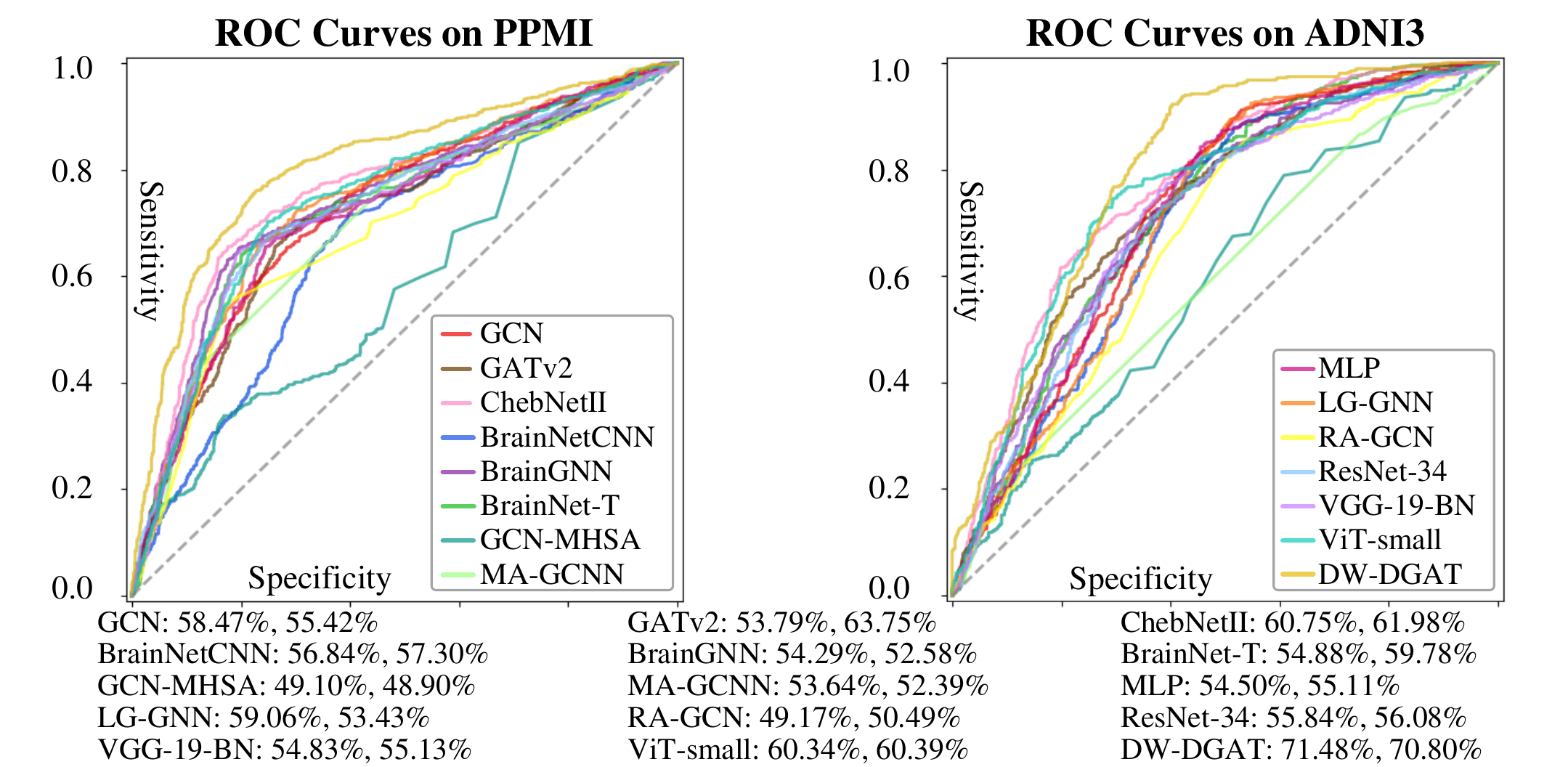}}
    \caption{ROC curves of all networks.}
    \label{fig3}
    \end{center}
\end{figure}

\begin{figure}[tb]
    \begin{center}
    \centerline{\includegraphics[width=\columnwidth]{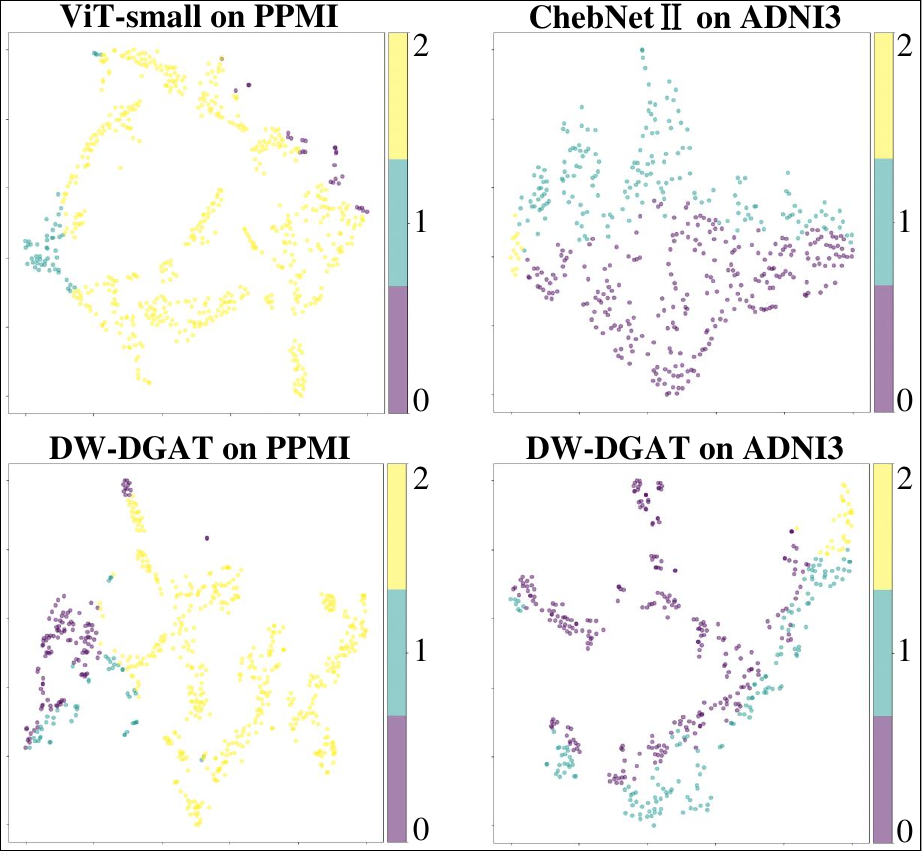}}
    \caption{The t-SNE visualization results of three networks on two datasets, where class-0 in purple indicates normal.}
    \label{fig4}
    \end{center}
\end{figure}

To visualize the AUCs of all compared networks, we plotted their ROC curves. As shown in Figure~\ref{fig3}, our ROC curves consistently dominate all others. Figure~\ref{fig4} presents the t-SNE visualizations of ViT-small, ChebNetII, and DW-DGAT. While DW-DGAT exhibits relatively less distinct inter-class boundaries, it achieves the most compact intra-class clusters, and identifies the largest number of samples from the minority classes.

\subsection{Ablation Experiments}
We conducted ablation experiments to investigate the contributions of each module in the proposed methodology, as shown in Table \ref{tab2}. Notably, the most contributive module is GGA, which improves accuracies by 4.25\% on HC \textit{vs.} PRO \textit{vs.} PD and 3.67\% on CN \textit{vs.} EMCI \textit{vs.} AD, totaling 7.92\%. The second most contributive module is DF, which improves accuracies by 2.36\% and 4.75\% on the two tasks, respectively, totaling 7.11\%. We attribute the success of GGA to its graph construction method and the MHSA-GC layers, which jointly enhance the relevance between phenotypic features and sample labels, enabling more precise message propagation. Specifically, the cosine similarity in graph construction mitigates the effects of large range variances in phenotypic features and normalizes the feature range to [0, 1]. This ensures that the $\sigma$ does not deviate excessively from the extreme values on either end, providing a balanced compromise between extremes \cite{gretton2006kernel}. Moreover, since the phenotypic features may not strictly align with the underlying sample relationships, the MHSA-GC layers adaptively adjust these relationships during neighbor aggregation, yielding more accurate neighborhood structures. In summary, the progressively increasing evaluation scores demonstrate the effectiveness of the proposed modules.

\section{Discussion}
\subsection{Temporal and Spatial Complexity Analysis}
Based on the notations defined in the Methodology section, we analyze the temporal and spatial complexities of the prediction computations in the DW-DGAT classifier, which are crucial considerations for clinical applications.

\subsubsection{Temporal Complexity Analysis.}
The temporal complexity of the embedding layer is $O(N \cdot R \cdot F' \cdot E)$. The temporal complexities of the MHSA and the feed-forward (FF) layer in the ViT encoder are $O(N \cdot (R+1)^2 \cdot E)$ and $O(N \cdot (R+1) \cdot E^2)$, respectively. Therefore, for $L=12$ encoder layers, the temporal complexity of the SGA is $O\left(N \cdot E \cdot (R \cdot F' + (R+1)^2 + (R+1) \cdot E) \right)$. The temporal complexities of the first and the second MHSA-GC layers are $O(N^3 \cdot E \cdot 2E)$ and $O(N^3 \cdot 2E \cdot 4E)$, respectively. The temporal complexity of the last linear layer in the GGA is $O(N \cdot 4E \cdot 2E)$. Thus, the temporal complexity of the GGA is $O(N^3 \cdot E^2)$. In total, the temporal complexity is $O(N \cdot E \cdot (R \cdot F' + (R+1)^2 + (R+1 + N^2) \cdot E)$. While the total complexity contains an $O(N^3)$ term, it reduces to $O(E \cdot (R \cdot F' + (R+1)^2 + (R+2) \cdot E)) = O(E \cdot R \cdot (F' + R + E))$ when the batch size $N$ is fixed. Under the reported experimental settings, the classifier requires 139.02 GFLOPs, while the generator requires 169.29 GFLOPs.

\subsubsection{Spatial Complexity Analysis.}
The spatial complexity of the embedding layer is $O(N \cdot R \cdot (F'+E))$. The spatial complexities of the MHSA and the FF layer are $O(N \cdot H' \cdot (1+R) \cdot (1+R+E))$ and $O(N \cdot (1+R) \cdot E)$, respectively, where $H'$ is the number of attention heads of the ViT encoder. Hence, for $L$ encoder layers and $F' < R \ll E$, the spatial complexity of the SGA is $O(N \cdot H' \cdot (1+R) \cdot (1+R+E))$. The spatial complexities of the first and the second MHSA GC layers are $O(N \cdot H \cdot N \cdot (N+E))$ and $O(N \cdot H \cdot N \cdot (N+2E))$, respectively. The spatial complexity of the last FC layer in the GGA is $O(N \cdot 2E)$. Thus, the spatial complexity of the GGA is $O(N^2 \cdot H \cdot (N+2E))$. Since $H < H' \le 2H$, the spatial complexity is $O(N \cdot H \cdot ((1+R) \cdot (1+R+E) + N \cdot (N + 2E)))$ in total. While the total complexity contains an $O(N^3)$ item, it reduces to $O(H \cdot ((1+R) \cdot (1+R+E) + 2E)) = O(H \cdot R \cdot (R + E))$ when the batch size $N$ is fixed. Under the reported experimental settings, the classifier consumes approximately 2728 MB of GPU memory, whereas the generator consumes approximately 3288 MB.

\section{Conclusion}
In this paper, we proposed a data fusion strategy to integrate multi-structured, high-dimensional metrics, providing a solid foundation for subsequent feature extraction. Next, we introduced DW-DGAT, a dual graph attention network with a class weight generator for ND diagnosis, which effectively captures reduced ROI graph features and fuses sample phenotypic information at both micro and macro levels, while mitigating class imbalance. A series of comparative and ablation experiments based on two datasets coming from the PPMI and the ADNI3 data sources demonstrated the superiority and generalizability of our proposed method, particularly the GGA module.

Nevertheless, our network is not without limitations. The 3D data fusion strategy does not account for the large differences in voxel numbers across ROIs, potentially neglecting latent features in larger ROIs.

\section{Acknowledgments}
This work was supported partly by National Natural Science Foundation of China (No. 62276172, 62306106), National Natural Science Foundation of Guangdong Province (No. 2023A1515011378), Shenzhen Key Basic Research Project (No. JCYJ20230808105602005), and China Scholarship Council.

\bibliography{aaai2026}

\appendix

\setcounter{figure}{0}
\setcounter{table}{0}
\renewcommand{\thetable}{A\arabic{table}}
\renewcommand{\thefigure}{A\arabic{figure}}

\section{Supplementary Material}

\subsection{Abstract}
In this supplementary material, we first introduce the process of obtaining data, and the PANDA settings for data preprocessing. Then, we complement principles for the data fusion strategy, and explain why it is general-purpose. Finally, we offer an in-depth analysis of the success of the adjacency graph construction.

\subsection{Download and Organize Data}
In this section, we delineate how to acquire MRI, DTI, and phenotypic data for our study.

Regarding the PPMI dataset, we accessed the data portal hosted by the Laboratory of Neuro-Imaging (LONI) (\url{https://ida.loni.usc.edu/pages/access/search.jsp?project=PPMI&tab=advSearch}) to download DICOM images from the \texttt{"Advanced Image Search"} page and CSV files from the \texttt{"Study Data"} page.

For DTI data, we selected the \texttt{"DTI"} and \texttt{"MRI"} fields and applied the \texttt{"AND"} logical connector in the \texttt{"IMAGE"} group to filter subjects who had underwent both MRI and DTI scans. To minimize the impact of scanner and acquisition parameter differences, we entered \texttt{"DTI\_gated"} in the \texttt{"Image Description"} textbox. Although the acquisition parameters still varied, all scans were performed using SIEMENS devices. The filtered DTI data was then downloaded.

For MRI data, we reused the subject IDs from the selected DTI scans by entering them into the \texttt{"Subject ID"} textbox, selected \texttt{"MRI"} in the \texttt{"Modality"} field, and checked the \texttt{"T1"} option under the \texttt{"Weighting"} field in the \texttt{"IMAGING PROTOCOL"} group. After downloading the MRI data, we obtained paired 3D scans for both structured MRI (sMRI) and DTI.

For phenotypic data, we downloaded the \texttt{"Curated Data Cuts Files"} from the \texttt{"Curated Data Cuts"} group of the \texttt{"Study Data"} page. Afterwards, we merged \texttt{"PPMI Curated Data Cut Public.csv"} and \texttt{"PPMI Original Cohort BL to Year 5 Dataset.csv"} based on common fields to maximize the sample count.

Regarding the ADNI3 dataset, its acquiring procedure was similar to that of PPMI. However, the link to the ADNI data portal is \url{https://ida.loni.usc.edu/pages/access/search.jsp?project=ADNI&tab=advSearch&page=SEARCH}. The \texttt{"ADNI3"} checkbox under the \texttt{"PROJECT/PHASE"} group was checked. In addition, as ADNI3 provides only a single CSV file, merging was not required.

The demographics of PPMI participants and their corresponding neuroimaging samples are presented in Table~\ref{appendix:tab1}, while those for ADNI3 are shown in Table~\ref{appendix:tab2}.

\begin{table}[th]
    \begin{center}
        \begin{small}
                \begin{tabular}{lcccc}
                \toprule
                    Cohort & HC & PRO & PD & Total  \\
                \midrule
                    Sex (M/F) & 45/24 & 37/35 & 113/62 & 195/121 \\
                    Age (A±S) & 61.3±10.8 & 62.9±8.2 & 62.1±9.3 & 62.1±9.4 \\
                    Subjects & 69 & 72 & 175 & 316 \\ \hline
                    0th month & 59 & 68 & 146 & 273 \\
                    12th month & 54 & 17 & 122 & 193 \\
                    24th month & 8 & 38 & 124 & 170 \\
                    Samples & 121 & 123 & 392 & 636 \\
                \bottomrule
                \end{tabular}
        \end{small}
    \end{center}
    \caption{PPMI subjects and neuroimaging samples. The Age row reports the averages and standard deviations.}
    \label{appendix:tab1}
\end{table}

\begin{table}[th]
    \begin{center}
        \begin{small}
                \begin{tabular}{lcccc}
                \toprule
                    Cohort & CN & EMCI & AD & Total  \\
                \midrule
                    Sex (M/F) & 75/88 & 65/54 & 17/12 & 157/154 \\
                    Age (A±S) & 71.6±6.2 & 70.2±6.9 & 74.0±7.8 & 71.3±6.7 \\
                    Subjects & 163 & 118 & 29 & 310 \\ \hline
                    0th month & 94 & 67 & 2 & 163 \\
                    12th month & 46 & 74 & 24 & 144 \\
                    24th month & 94 & 52 & 11 & 157 \\
                    Samples & 234 & 193 & 37 & 464 \\
                \bottomrule
                \end{tabular}
        \end{small}
    \end{center}
    \caption{ADNI3 subjects and neuroimaging samples. The Age row reports the averages and standard deviations.}
    \label{appendix:tab2}
\end{table}

\subsection{PANDA Settings for Data Preprocess}

\begin{figure*}[t]
    \centering
    \includegraphics[width=\textwidth]{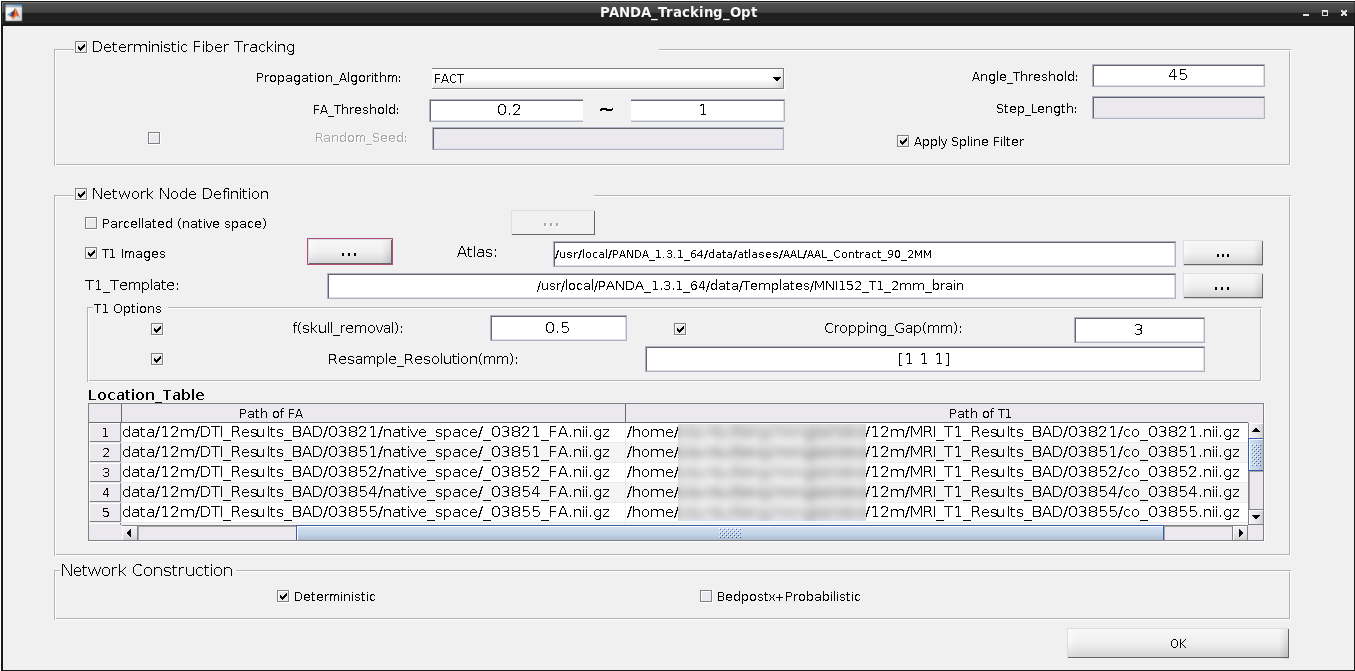}
    \caption{PANDA tracking options.}
    \label{appendix:fig3}
\end{figure*}

\begin{figure}[t]
    \centering
    \includegraphics[width=0.9\columnwidth]{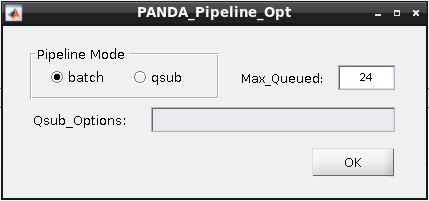}
    \caption{PANDA pipeline options.}
    \label{appendix:fig1}
\end{figure}

\begin{figure}[t]
    \centering
    \includegraphics[width=\columnwidth]{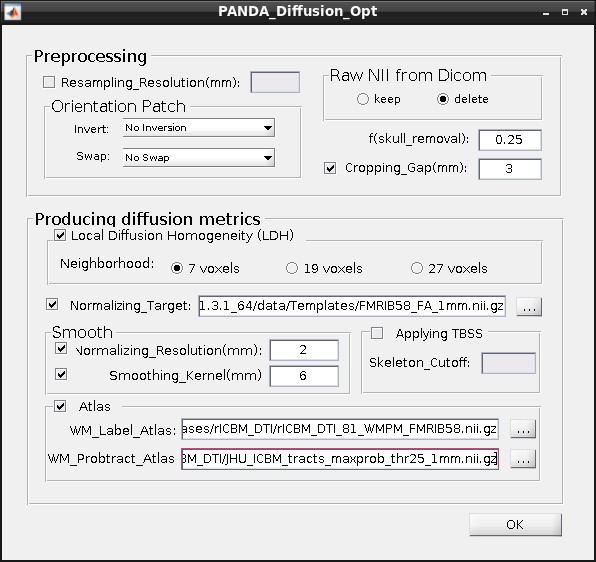}
    \caption{PANDA diffusion options.}
    \label{appendix:fig2}
\end{figure}

\begin{figure}[t]
    \centering
    \includegraphics[width=\columnwidth]{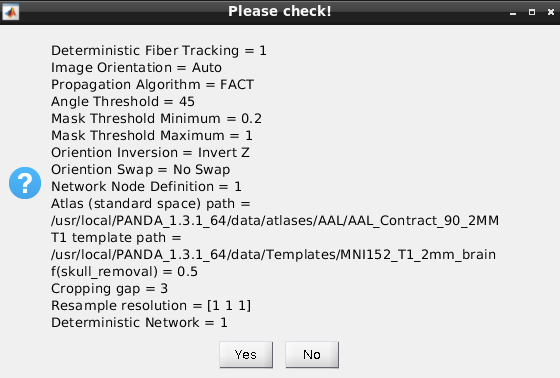}
    \caption{A summary of all running arguments of PANDA.}
    \label{appendix:fig4}
\end{figure}

Data preprocessing was performed using MATLAB 2018a or 2019a on Ubuntu 18.04. The versions of FSL and PANDA were 6.0.7.13 and 1.3.1 (x64), respectively. We first used the \texttt{"DICON->NIfTI"} PANDA utitlity to transform the data format of MRI data. The \texttt{"Full Pipeline"} of PANDA utilities was then employed to preprocess both MRI and DTI data. The configuration includes pipeline options, diffusion options, and tracking options, which are illustrated in Figure~\ref{appendix:fig1}, Figure~\ref{appendix:fig2}, and Figure~\ref{appendix:fig3}, respectively. Figure~\ref{appendix:fig4} summarizes all runtime parameters. After preprocessing, we conducted quality control by inspecting the images in the \texttt{"quality\_control"} folder following the PANDA manual \cite{cuipanda}, and discarded samples of poor quality.

\subsection{Principles of Data Fusion}
We describe the principles of data fusion adopted in our framework.

For the two 1D vectors, we apply element-wise division because surface size quantifies the number of voxels where fibers terminate within an ROI, whereas voxel size represents the total number of voxels in the ROI \cite{cuipanda}.

The rationale for processing the 2D deterministic networks is threefold: (a) the three types of network data vary significantly in range, necessitating min-max normalization; (b) the brain networks are inherently sparse \cite{wee2014group}, and applying the $L_1$-norm helps avoid generating excessively large or small values that could cause gradient explosion or vanishing; and (c) parkinsonian symptoms caused by lesions in brain regions such as the globus pallidus internus or subthalamic nucleus may be compensated by additional connections, such as the \texttt{"}hyper-direct pathway\texttt{"} \cite{poewe2017parkinson}, which can affect the brain network\textquotesingle{}s row sums.

\begin{table*}[t]
    \begin{center}
    \begin{small}
    \begin{tabular}{lcccccccc}
        \toprule
        \multirow{2}{*}[-0.5ex]{\makecell[l]{Graphs}} & \multicolumn{4}{c}{\textbf{HC \textit{vs.} PRO \textit{vs.} PD}} & \multicolumn{4}{c}{\textbf{NC \textit{vs.} EMCI \textit{vs.} AD}} \\
        \cmidrule(lr){2-5} \cmidrule(lr){6-9}
        ~ & ACC(\%) & BA(\%) & F1(\%) & SPE(\%) & ACC(\%) & BA(\%) & F1(\%) & SPE(\%) \\
        \hline
        Euclidean & 62.53±4.15 & 35.39±2.20 & 49.34±4.86 & 67.70±1.10 & 61.95±3.33 & 43.93±2.17 & 57.26±3.48 & 71.96±1.08 \\
        Phenotype & 71.70±5.30 & 55.59±9.58 & 68.42±10.16 & 77.79±4.79 & 65.09±3.14 & 64.31±9.97 & 64.66±10.44 & 82.15±4.99 \\
        \textbf{Relationship} & \textbf{94.03±7.58} & \textbf{89.10±14.07} & \textbf{92.43±9.98} & \textbf{94.55±7.03}& \textbf{100.00±0.00} & \textbf{100.00±0.00}& \textbf{100.00±0.00} & \textbf{100.00±0.00} \\
        \bottomrule
    \end{tabular}
    \end{small}
    \end{center}
    \caption{Comparison of three types of adjacency graphs on two classification tasks.}
    \label{appendix:tab3}
\end{table*}

In addition, we argue that our data fusion strategy is general-purpose, as it operates based on the data’s dimensions rather than specific data types. Many neurological datasets can be categorized as 1D (\textit{e.g.}, regional statistics), 2D (\textit{e.g.}, ROI-wise connectivity networks), or 3D (\textit{e.g.}, neuroimaging volumes).

For 1D data, we use division to fuse features based on their semantic significance. Alternatively, addition or multiplication could be applied if the vectors are positively correlated. Even when uncorrelated, vectors can still be stacked to form a matrix for further use.

For 2D data, prior studies suggest that the human brain exhibits a compensatory mechanism, wherein other regions may strengthen their connectivity when certain areas are impaired \cite{voytek2010dynamic}—a common pattern across multiple neurological disorders.

For 3D neuroimages, features such as centroid coordinates, centroid intensity, average intensity, and maximum intensity are widely available. Thus, our fusion approach is broadly applicable to a range of neurological studies, including autism spectrum disorder (ASD), obsessive-compulsive disorder (OCD), infant brain age analysis, and others.

\begin{figure*}[t]
    \begin{center}
        \centerline{\includegraphics[width=\textwidth]{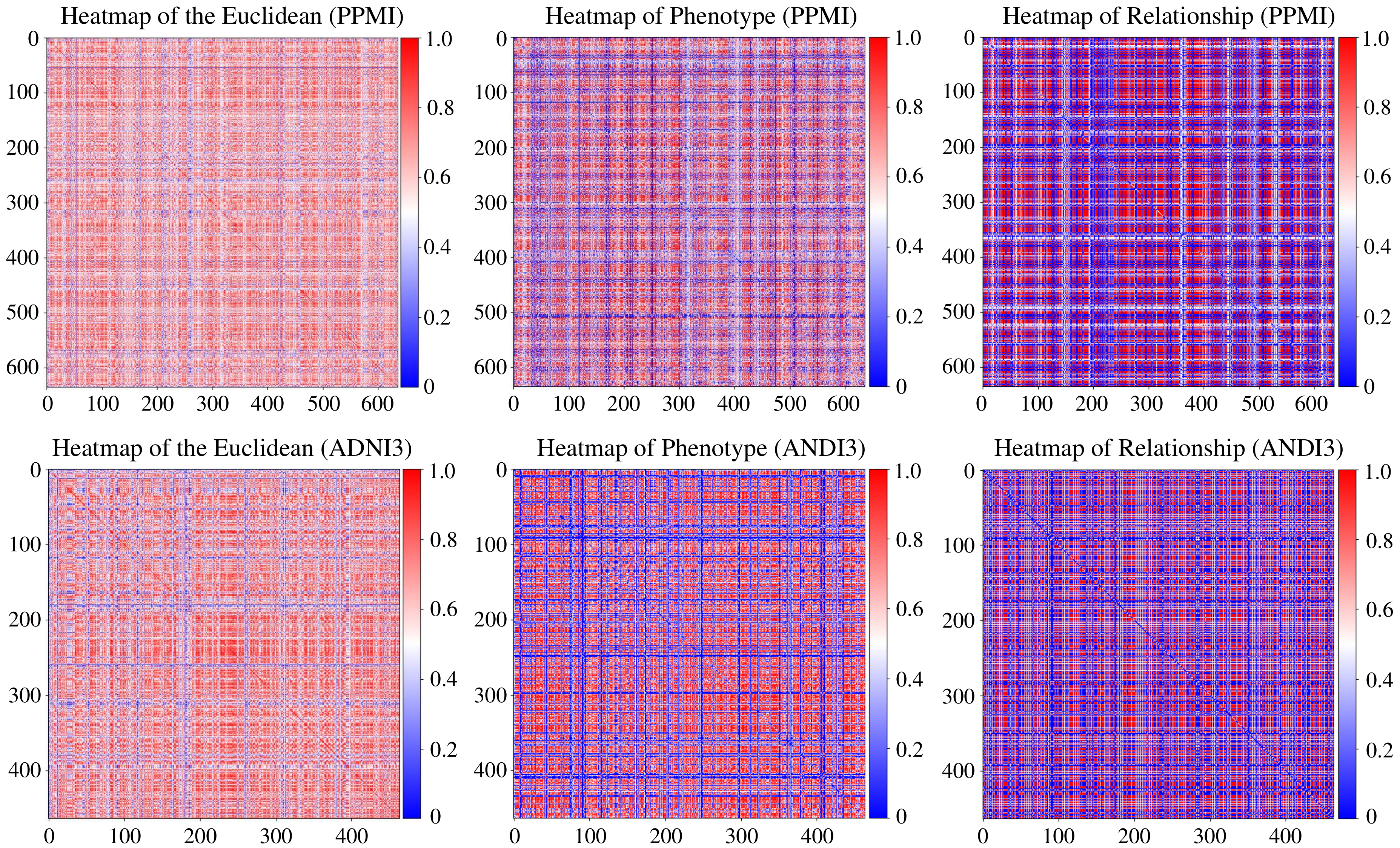}}
        \caption{Heatmaps of different types of adjacency graphs on two datasets.}
        \label{appendix:fig5}
    \end{center}
\end{figure*}

\begin{figure*}[t]
    \begin{center}
        \centerline{\includegraphics[width=\textwidth]{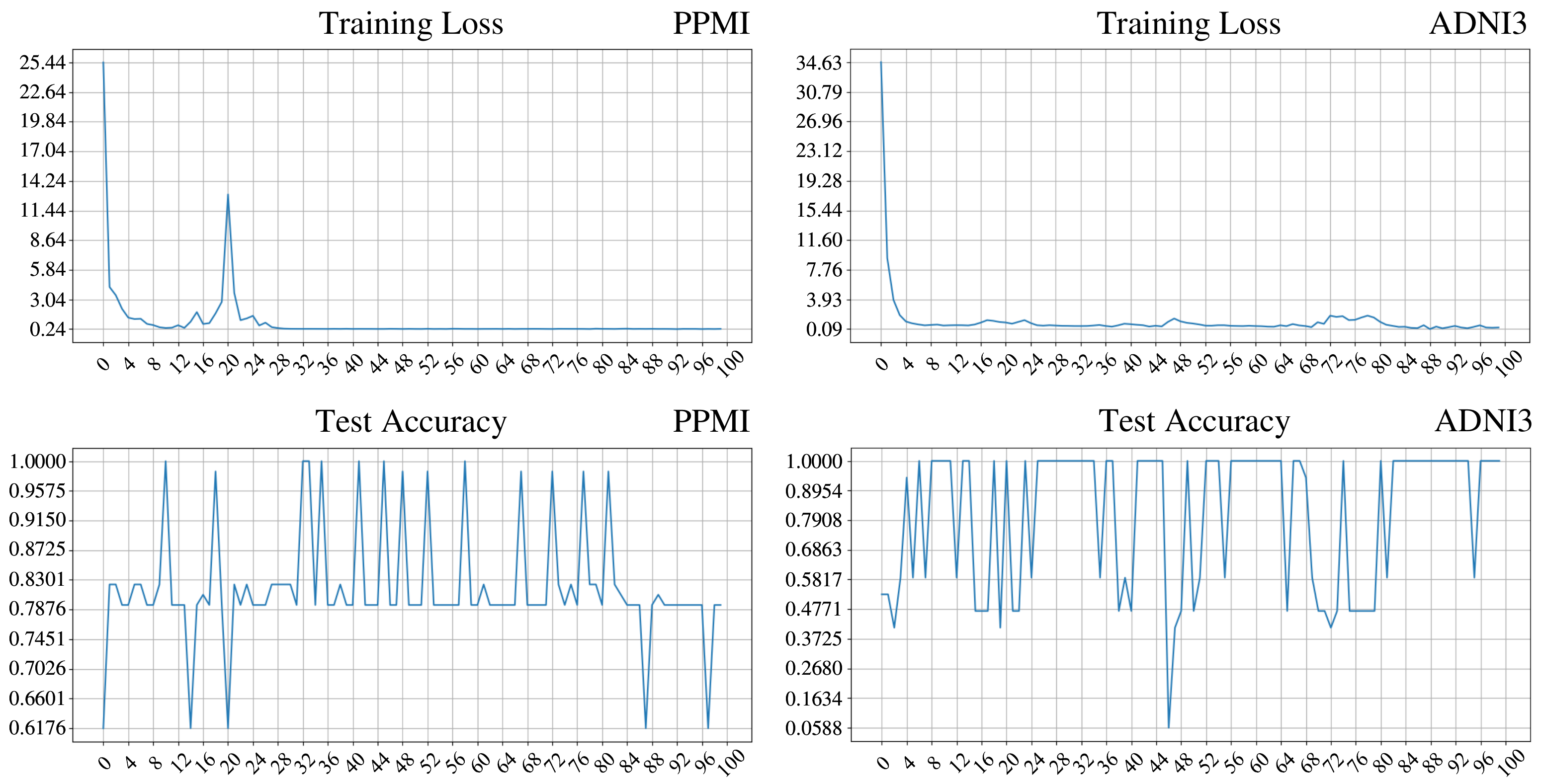}}
        \caption{Histories of training loss and test accuracy over 100 epochs on two datasets.}
        \label{appendix:fig6}
    \end{center}
\end{figure*}

\subsection{Analysis of Adjacency Graph Construction}
As mentioned in the Global Graph Attention subsection, an adjacency graph where node relationships closely reflect the true relationships among ground truth samples can substantially improve the classification accuracy. To demonstrate this, we compared the performance between the Euclidean graph and the phenotypic graph on our network that integrates the DF, SGA, and GGA modules, along with a relationship graph solely based on the GGA module that uses only the FA deterministic brain networks, as shown in Table~\ref{appendix:tab3}. The Euclidean graph is constructed by computing pairwise distances between nodes based on sample features, thus excluding phenotypic data. The phenotypic graph is the $\mathbf{A}_1$ described in the Global Graph Attention subsection. The relationship graph is derived from sample labels: the weight of an edge is set to 1 if two nodes share the same class label, and 0 otherwise. Both the Euclidean and phenotypic networks were trained for 500 epochs, while the relationship network was trained for only 100 epochs. All networks were evaluated via ten-fold cross-validation. From Table \ref{appendix:tab3}, we observe that the phenotypic graph leads to better performance than the Euclidean graph, while the relationship graph achieves the highest classification accuracies, even reaching perfect classification on the ADNI3 dataset.

To further explore these differences, Figure~\ref{appendix:fig5} visualizes the heatmaps of the three graphs on both datasets. The phenotypic graphs more closely resemble the ground-truth relationship graphs, whereas the Euclidean graphs exhibit values fluctuating within a narrow range and are often indistinguishable, which hampers their ability to aggregate meaningful neighbor information. This observation is consistent with the performance gap between the Euclidean and phenotypic networks.

Finally, to highlight the importance of accurate adjacency graph construction, we plot the training loss and test accuracy curves of the relationship network on the first validation fold, as shown in Figure~\ref{appendix:fig6}. The network achieves 100\% accuracy in fewer than 50 epochs, further validating the efficacy of well-constructed adjacency graphs.


\end{document}